\documentclass[conference]{IEEEtran}
\IEEEoverridecommandlockouts
\usepackage{cite}
\usepackage{amsmath,amssymb,amsfonts}
\usepackage{algorithmic}
\usepackage{graphicx}
\usepackage{textcomp}
\usepackage[nointegrals]{wasysym}
\DeclareUnicodeCharacter{266B}{\twonotes}
\usepackage{xcolor}
\usepackage{enumitem}
\usepackage{hyperref}
\hypersetup{hidelinks}
\usepackage{booktabs}
\usepackage{multirow}
\usepackage{makecell}
\newcommand{\best}[1]{\textbf{#1}}
\newcommand{\second}[1]{\underline{#1}}
\def\BibTeX{{\rm B\kern-.05em{\sc i\kern-.025em b}\kern-.08em
    T\kern-.1667em\lower.7ex\hbox{E}\kern-.125emX}}
\begin{document}

\title{Parameters vs. Context: TRACE Fine-Tuning for Robust Retrieval-Augmented Generation}

\author{
\IEEEauthorblockN{
Zhengchen Huang\textsuperscript{1},
Yundong Sun\textsuperscript{1}\textsuperscript{,*}
,
Minrui Song\textsuperscript{2},
Shuanglong Yao\textsuperscript{1},
Ye Liu\textsuperscript{1},
Ji Chen\textsuperscript{1},
Xing Wang\textsuperscript{1}
}

\IEEEauthorblockA{
\textsuperscript{1}
School of Information Science \& Engineering, LinYi University, China
}

\IEEEauthorblockA{
\textsuperscript{2}
School of Computer Science and Technology, Harbin Institute of Technology, China
}

\IEEEauthorblockA{
\textsuperscript{*}Corresponding author: Yundong Sun, hitffmy@163.com
}
}

\bibliographystyle{IEEEtran}
\maketitle

\begin{abstract}

Retrieval-Augmented Generation (RAG) mitigates knowledge obsolescence and factual hallucination in large language models by introducing external context. However, when retrieved knowledge conflicts with the model’s internal parametric knowledge, the model may either blindly follow misleading context or incorrectly rely on parametric knowledge, leading to unreliable responses. To address this issue, this paper proposes TRACE (Debate-\underline{TR}ace and \underline{A}nswer-\underline{C}ompleteness r\underline{E}gularized fine-tuning), a robust fine-tuning framework for RAG under knowledge conflicts. First, we propose a fine-tuning method that leverages multi-agent debate traces to extract correct candidates, incorrect candidates, and answer-shift patterns, providing fine-grained supervision for reliable knowledge-source selection. In addition, we design an answer completeness regularization mechanism to alleviate empty, overly short, and prematurely terminated responses via answer-tail token reinforcement and premature termination suppression. The fine-tuning objective combines correct-answer supervision, incorrect-candidate suppression, answer-tail token reinforcement, and premature termination suppression, enabling the model to use reliable external context, resist misleading or irrelevant retrieved content, and fall back to parametric knowledge when retrieved evidence is unreliable. Experiments across multiple knowledge-conflict scenarios and datasets show that TRACE improves robustness against misleading retrieved knowledge and reduces incomplete answers. These results demonstrate that multi-agent debate traces and answer completeness regularization jointly enhance knowledge-source selection, conflict robustness, and answer quality in RAG models. Our code is available at \href{https://github.com/PHD-lanyu/TRACE}{TRACE}.

\end{abstract}


\section{Introduction}
Large Language Models (LLMs), such as GPT~\cite{singh2025openai} and Llama~\cite{touvron2023llama}, have become strong backbones for open-domain question answering, knowledge reasoning, and text generation. Despite this progress, their responses still depend on parametric knowledge that can be outdated, incomplete, or incorrect~\cite{lin2022truthfulqa}, which may lead to factual hallucinations~\cite{ji2023survey} and unreliable answers. Retrieval-Augmented Generation (RAG)~\cite{lewis2020retrieval} mitigates this problem by conditioning generation on retrieved documents or knowledge-base content, thereby allowing models to use external evidence beyond their fixed parameters~\cite{gao2023retrieval,guu2020retrieval}.

\begin{figure}[t]
\centerline{\includegraphics[width=0.88\linewidth,trim=0 20 0 0]{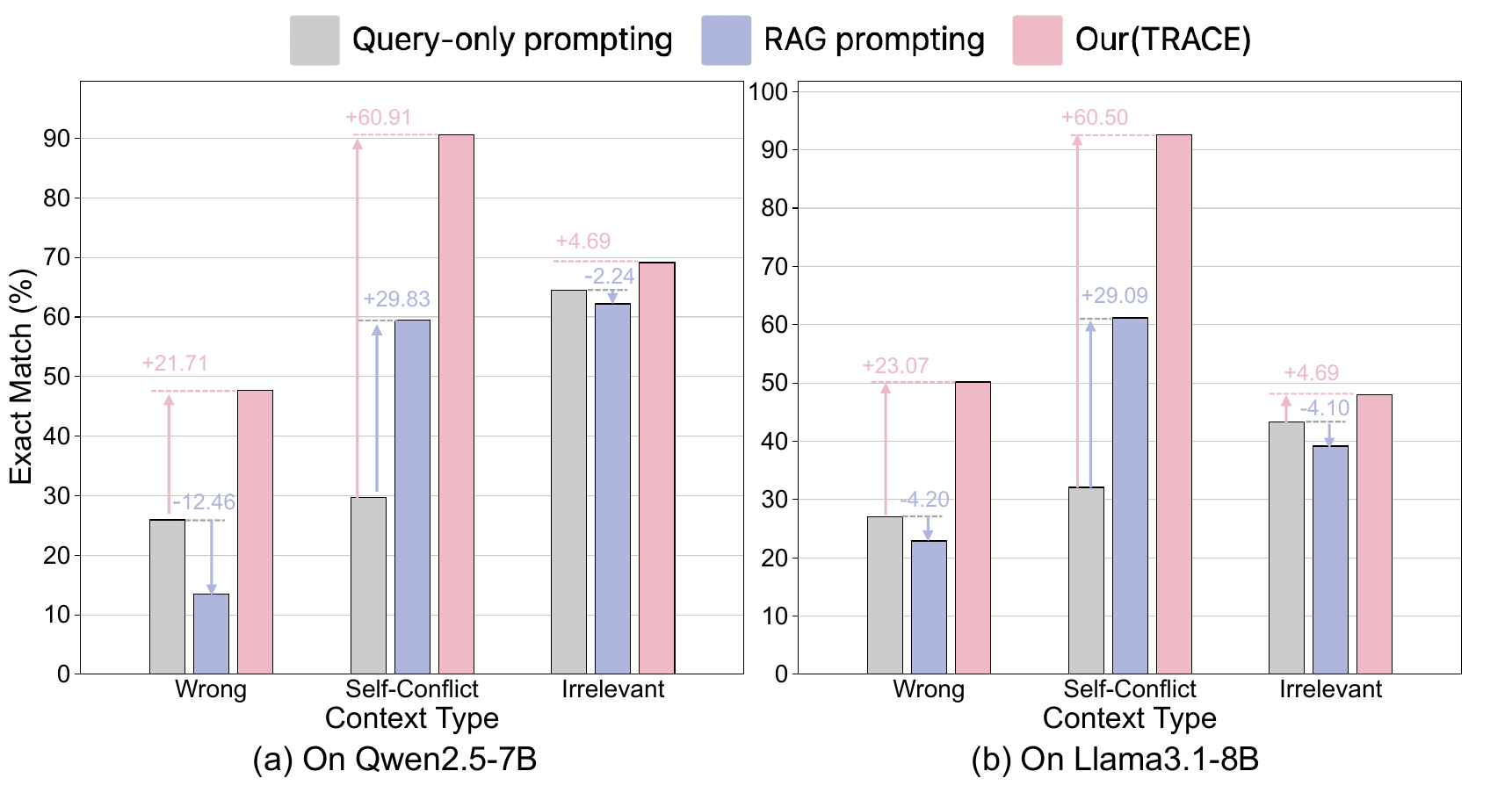}}
\caption{Performance under different knowledge-conflict contexts on ConFiQA-MR, ConFiQA-SC, and ExplainPE.}
\label{fig:motivation}
\vspace{-0.5cm}
\end{figure}

The central difficulty in RAG is that retrieved context can conflict with the model's parametric knowledge. In practical retrieval pipelines, external context may be correct, misleading, self-conflicting, or irrelevant~\cite{longpre2021entity}. As shown in Fig.~\ref{fig:motivation}, conventional RAG prompting becomes unstable when the retrieved context is unreliable: incorrect or irrelevant context can mislead the model and even make it perform worse than query-only prompting, while self-conflicting context provides only limited gains. These observations indicate that LLMs do not inherently know when to trust retrieved evidence, when to rely on parametric knowledge, or how to revise an answer when the two sources disagree~\cite{xie2024adaptive}. Therefore, robust RAG requires source-aware generation: the model must use reliable retrieved evidence, resist misleading context, and preserve correct parametric knowledge when retrieval is unreliable.

Existing work improves RAG under knowledge conflicts from three main directions: prompt-based knowledge integration~\cite{wang2025astute}, retrieval-quality assessment~\cite{yan2024corrective}, and learnable knowledge-selection strategies~\cite{bi2026parameters}. Prompt-based methods guide models to compare internal knowledge with external context~\cite{wei2025instructrag}; retrieval-quality assessment methods identify, filter, or correct low-quality retrieved content~\cite{xiang2024certifiably}; and learnable selection methods use decoding control, preference optimization, or reinforcement learning to choose among information sources under conflicting conditions~\cite{choi2025conflict}. However, these methods still face two limitations. First, many methods introduce additional prompts, reflection stages, retrieval evaluation modules, or multi-step reasoning at test time, which increases inference overhead. Second, their training signals are usually constructed from final answers, decoding distributions, or sampled rewards, so they rarely preserve the intermediate candidate answers, erroneous paths, and answer revisions that appear during conflict reasoning. As a result, the supervision is often too coarse to teach how a model should move from a misleading answer to a reliable one.

Multi-agent debate provides a natural source of process-level supervision for this problem. Prior studies show that interaction, questioning, and revision among multiple model instances can improve factual judgment and complex reasoning~\cite{choi2026debate,du2024improving}. Inspired by this observation, we propose TRACE (Debate-\underline{TR}ace and \underline{A}nswer-\underline{C}ompleteness r\underline{E}gularized fine-tuning), a robust fine-tuning method for RAG under knowledge conflicts. TRACE uses multi-agent debate only during training, rather than retaining debate at inference time. It extracts correct candidate answers, incorrect candidate answers, and answer-shift traces from debate trajectories, then converts these traces into positive and negative fine-tuning signals. In this way, the model learns not only which answer is correct, but also which misleading candidates should be suppressed and how answer revisions occur when parametric and retrieved knowledge conflict.

Reliable source selection alone is not sufficient, because knowledge-conflict fine-tuning can also introduce answer-incompleteness errors. In our empirical analysis, some models identify the correct answer direction but stop too early, producing empty answers, overly short answers, incomplete entities, or unclear answer boundaries. TRACE therefore adds answer completeness regularization, which reinforces answer-tail tokens and suppresses premature termination inside the answer span. This auxiliary constraint encourages the model to produce complete and parsable final answers after selecting the appropriate knowledge source.

The main contributions of this paper are as follows:

\begin{enumerate}[label=(\arabic*)]
\item We propose a debate-trace fine-tuning method that converts correct candidates, incorrect candidates, and answer-shift traces into fine-grained supervision signals, enabling the model to select more reliable knowledge when parametric and retrieved knowledge conflict.

\item We identify an answer-incompleteness issue that can emerge during knowledge-conflict fine-tuning, and design answer completeness regularization to reduce empty, overly short, and incomplete outputs through answer-tail token reinforcement and premature termination suppression.

\item We conduct experiments across correct, wrong, self-conflicting, irrelevant, and partially irrelevant retrieval scenarios. The results show that TRACE improves robustness in explicit conflict settings while preserving the use of correct retrieved knowledge, and ablation studies validate the proposed debate-trace supervision and answer completeness regularization modules.
\end{enumerate}

\section{RELATED WORK}
\subsection{RAG under Knowledge Conflicts}
RAG under knowledge conflicts aims to decide when retrieved context should override, complement, or be ignored in favor of parametric knowledge. Prompt-based methods address this problem by explicitly eliciting source comparison or self-critique during inference. Astute RAG~\cite{wang2025astute} elicits internal knowledge and integrates it with retrieved content in a source-aware manner, Self-RAG~\cite{asai2024self} controls retrieval and critique through reflection tokens, and InstructRAG~\cite{wei2025instructrag} uses self-generated rationales to extract valid evidence from noisy retrieval. These methods make the reasoning process more interpretable, but they typically depend on additional prompts, reflection steps, or multi-stage reasoning at test time.

Retrieval-quality assessment methods improve RAG by judging or repairing the external context before generation. CRAG~\cite{yan2024corrective} estimates retrieval reliability and invokes corrective retrieval when the context is unreliable, RobustRAG~\cite{xiang2024certifiably} studies robustness against retrieval corruption, and TruthfulRAG~\cite{liu2026truthfulrag} converts retrieved text into a knowledge graph to identify factual-level conflicts. These methods reduce the risk of misleading retrieval, but their main focus is obtaining or filtering better context rather than training the generator to choose between parametric and retrieved knowledge. In contrast, TRACE assumes that the retrieved context may remain correct, wrong, self-conflicting, or irrelevant, and trains the model to make the source-selection decision inside generation.

\subsection{Learnable Knowledge-Source Selection}

Learnable knowledge-selection methods are the closest line of work to TRACE because they directly optimize how a model uses parametric and contextual knowledge. CK-PlUG~\cite{bi2026parameters} controls reliance on parametric and contextual knowledge during decoding, Context-DPO~\cite{bi2026context} improves context faithfulness through preference pairs, KnowPO~\cite{zhang2025knowpo} formulates knowledge-aware preference optimization, and Knowledgeable-R1~\cite{lin2026resisting} strengthens resistance to contextual interference through parametric-knowledge reinforcement. These methods provide important baselines for source-aware RAG, but their supervision is mainly derived from final answers, sampled responses, reward signals, or decoding-level control. TRACE differs by using debate trajectories to preserve correct candidates, incorrect candidates, and answer shifts, so the fine-tuning data contains process-level evidence about how answers become wrong or get corrected under conflict.

Existing conflict-oriented RAG work also pays limited attention to answer completeness after source selection. Most prior methods evaluate whether the model follows the appropriate knowledge source or produces a factually correct final answer, whereas our observations show that knowledge-conflict fine-tuning can produce empty answers, overly short answers, incomplete entities, and unclear answer boundaries. TRACE therefore treats final-answer completeness as a separate generation-side objective, complementing source selection with answer-tail token reinforcement and premature-termination suppression.

\subsection{Multi-agent Debate as Process-level Supervision}

Multi-agent debate exposes intermediate errors, corrections, and answer revisions that are difficult to observe from a single final response. Liang et al.~\cite{liang2024encouraging} introduced a debate framework in which multiple agents discuss a problem and a judge determines the final answer. Subsequent studies show that debate can improve factual judgment and reasoning~\cite{du2024improving,choi2026debate}, support consensus formation among diverse models~\cite{chen2024reconcile}, and provide trajectories for post-training or preference optimization~\cite{motwani2025malt,zhou2025debate}. These findings suggest that debate records are not only inference-time reasoning traces, but also potential supervision sources.

The connection between debate traces and RAG knowledge conflicts remains underexplored. Knowledge conflicts are common in LLM applications~\cite{xu2024knowledge}, and recent work studies how to verify or select knowledge under inconsistent evidence~\cite{zeng2025towards}. However, existing RAG conflict methods usually do not mine debate trajectories for source labels, positive candidates, negative candidates, and answer-shift patterns. TRACE fills this gap by using multi-agent debate only during data construction, then converting the resulting trajectories into conflict-aware supervision for a single target model. This design preserves the training value of debate while avoiding debate-time overhead during deployment.

\section{METHODOLOGY}

\subsection{Task Definition}

Given a question $Q$ and a retrieved context $C$, a RAG model generates an answer $A=f_{\theta}(Q,C)$. The retrieved context may be reliable, misleading, irrelevant, or internally conflicting, and the model may also rely on its parametric knowledge. This paper focuses on answerable knowledge-conflict settings in which at least one source can support the golden answer. When parametric and retrieved knowledge disagree, the model should select the more reliable source; when both sources are reliable, it should maintain stable answer generation rather than over-suspecting either source.

\subsection{Overview}

\begin{figure*}[t]
\centerline{\includegraphics[width=0.82\linewidth,trim=0 10 0 30]{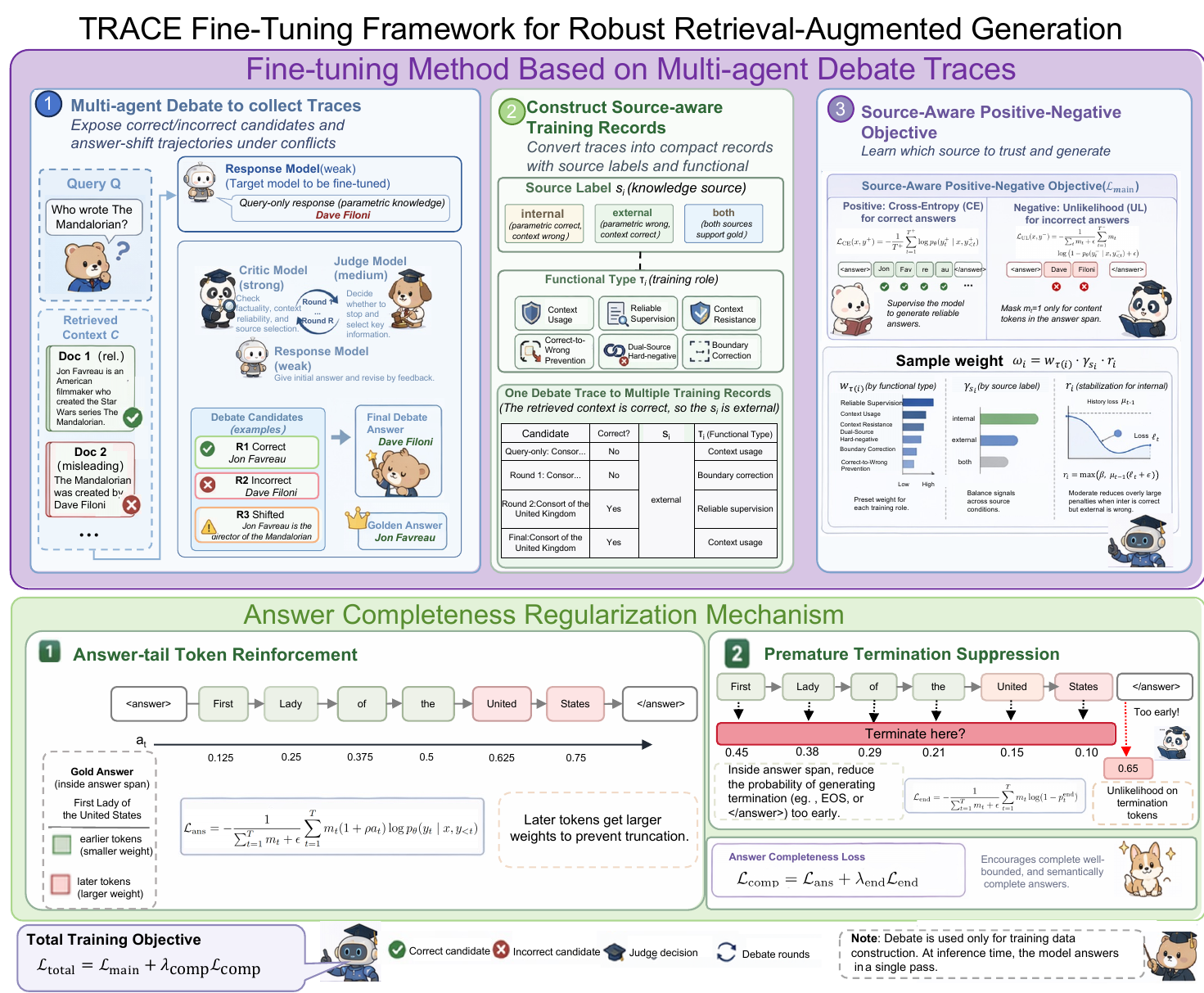}}
\centering
\caption{Overview of TRACE. Debate is used only during data construction to expose source-selection errors and answer shifts. The target model is then fine-tuned with debate-trace supervision and answer completeness regularization.}
\label{fig:flow_chart}
\vspace{-0.5cm}
\end{figure*}

TRACE addresses the two methodological gaps identified in the Introduction and Related Work: coarse conflict supervision and incomplete final answers after knowledge-conflict fine-tuning. As shown in Fig.~\ref{fig:flow_chart}, TRACE first uses multi-agent debate to expose correct candidates, incorrect candidates, and answer-shift trajectories under knowledge conflicts. It then converts these traces into source-aware positive and negative fine-tuning records, allowing the target model to learn which knowledge source to trust without running debate at inference time. Finally, TRACE adds answer completeness regularization to reduce empty, overly short, and prematurely terminated answers.

\subsection{Fine-tuning Method Based on Multi-agent Debate Traces}

The first component of TRACE constructs process-level supervision from multi-agent debate traces. This component contains three steps: collecting debate traces under retrieved context, converting the traces into source-labeled training records, and optimizing the target model with positive supervision for reliable answers and unlikelihood suppression for misleading candidates. This design turns debate from an inference-time procedure into a training-time supervision source.

\subsubsection{Debate Trace Collection}
For each training sample, TRACE first queries the base model with $Q$ only to obtain a query-only response, which provides an estimate of the model's parametric-knowledge behavior. TRACE then conducts a multi-round debate with the retrieved context $C$. The debate has three roles: the response model produces candidate answers, the critic model checks factual consistency, context reliability, and knowledge-source selection, and the judge model decides whether the debate should stop and selects key information from the final candidates.

TRACE uses a strong-to-weak debate setting to make the trace informative for fine-tuning. The response model is the weak target model to be fine-tuned, the critic model is stronger, and the judge model has intermediate capability. This configuration encourages the debate to expose misleading candidates, corrections, and answer shifts that are difficult to obtain from a single final response. Because debate is used only to construct training data, TRACE does not introduce multi-agent interaction or additional reasoning rounds during deployment.

\subsubsection{Construction of Source-Aware Training Records}
TRACE converts each debate trace into compact training records instead of directly using the full debate transcript as model input. Each record is built from the query-only response, intermediate debate candidates, the final debate answer, and the golden answer. This representation avoids overfitting to surface debate language and focuses the training signal on source selection, candidate correction, and answer stability.

Each training record receives a knowledge-source label $s_i \in \{\mathrm{internal}, \mathrm{external}, \mathrm{both}\}$. The \emph{internal} label indicates that parametric knowledge supports the golden answer while the retrieved context is unreliable; the \emph{external} label indicates that the retrieved context supports the golden answer while the query-only response is incorrect; and the \emph{both} label indicates that both sources support the golden answer. The first two labels correspond to conflict settings, while \emph{both} serves as a non-conflict anchor that preserves ordinary RAG answering ability. If neither source supports the golden answer, TRACE excludes the sample because it is closer to unanswerability detection than source-aware answer generation.

On top of source labels, TRACE assigns functional type labels according to each constructed record's training role. These roles include reliable supervision, context usage, context resistance, correct-to-wrong prevention, dual-source error suppression, and boundary correction. Intermediate and final debate answers are divided into correct and incorrect candidates according to the golden answer, with at most two representative incorrect candidates retained from each debate trace to reduce noisy negatives. Table~\ref{tab:sample_taxonomy} summarizes the sample taxonomy.

\begin{table*}[t]
\centering
\caption{Taxonomy of fine-tuning samples for TRACE.}
\label{tab:sample_taxonomy}
\scriptsize
\setlength{\tabcolsep}{3pt}
\renewcommand{\arraystretch}{1.12}
\begin{tabular}{p{0.12\textwidth}p{0.16\textwidth}p{0.34\textwidth}p{0.30\textwidth}}
\toprule
\textbf{Knowledge-source label} & \textbf{Functional type} & \textbf{Construction condition} & \textbf{Training role} \\
\midrule

\multirow[c]{3}{0.12\textwidth}{\raggedright   \newline \newline internal / external \newline/ both}
& Reliable supervision 
& Golden answers or reliable correct candidate answers observed during debate. 
& Supervises correct answers and preserves basic answering ability. \\

& Correct-to-wrong prevention 
& A correct candidate appears in an intermediate round, but the later or final response drifts to an incorrect answer. 
& Suppresses shifts from correct candidates to incorrect answers and improves answer stability. \\

& Boundary correction 
& The candidate is close to the correct answer but is overly short, boundary-unclear, or difficult to judge. 
& Improves answer completeness, parsability, and final evaluation determinability. \\

\cmidrule(lr){1-4}

external 
& Context usage 
& The retrieved context supports the golden answer, but the query-only response is incorrect. 
& Trains the model to actively use external evidence when it is reliable. \\
\addlinespace[2pt]

internal 
& Context resistance 
& The query-only response is correct, but the retrieved context contains misleading or irrelevant. 
& Trains the model to avoid blindly following erroneous retrieved content. \\
\addlinespace[2pt]

both 
& Dual-source hard negative 
& Both parametric knowledge and retrieved context support the correct answer, but incorrect candidates still appear during debate. 
& Reduces the probability of abnormal incorrect candidates under dual-source reliable conditions. \\

\bottomrule
\end{tabular}
\vspace{-0.4cm}
\end{table*}

A single debate trace can be expanded into multiple training records because different candidates in the same trace may serve different training roles. During training, the source label $s_i$ determines the source-balance coefficient $\gamma_{s_i}$, and the functional type label $\tau_i$ determines the fixed sample-type weight $w_{\tau(i)}$. The main tunable parameters in this component are the incorrect-candidate suppression weight $\lambda_{\mathrm{neg}}$ and the source-balance coefficients $\gamma_{\mathrm{int}}$, $\gamma_{\mathrm{ext}}$, and $\gamma_{\mathrm{both}}$.

\subsubsection{Source-Aware Positive-Negative Objective}
Let $x=(Q,C)$ denote the input, $y^+$ denote a correct answer, and $y^-$ denote an incorrect candidate answer. For correct answers, TRACE uses cross-entropy (CE) supervision:
\begin{equation}
\mathcal{L}_{\mathrm{CE}}(x,y^+)=-\frac{1}{T^+}\sum_{t=1}^{T^+}\log p_{\theta}(y_t^+\mid x,y_{<t}^+),
\label{eq:ce_loss}
\end{equation}
where $T^+$ is the length of the correct answer. This term teaches the model to generate reliable answers under each source condition.

For incorrect candidates, TRACE uses an unlikelihood (UL) objective~\cite{Welleck2020Neural} to reduce the probability of misleading answer content. To avoid damaging the output template, this loss is computed only inside the answer span:
\begin{equation}
\begin{aligned}
\mathcal{L}_{\mathrm{UL}}(x,y^-)
&=-\frac{1}{\sum_{t}m_t+\epsilon}
\sum_{t=1}^{T^-}m_t \\
&\quad \log\left(1-p_{\theta}(y_t^-\mid x,y_{<t}^-)+\epsilon\right),
\end{aligned}
\label{eq:ul_loss}
\end{equation}
where $m_t\in\{0,1\}$ is the answer-content mask. The mask equals 1 only for tokens inside the answer span, so structural tags and answer-boundary tokens are not treated as negative samples. This design suppresses wrong answer content without encouraging malformed or incomplete answer boundaries.
{\parskip=0pt\par\indent
At the same time, TRACE also uses lightweight balance modulation as an auxiliary stabilization strategy for cases where parametric knowledge is correct but external retrieval is wrong. This component follows the motivation of Knowledge Balance Modulation in Knowledgeable-R1~\cite{lin2026resisting}, but it is not treated as the main contribution of this paper.}
{\parskip=0pt\par\indent
During the early stage of fine-tuning, the loss can be large and later gradually decreases. Therefore, we maintain an exponential moving average (EMA) $\mu_t$ to represent the historical loss level of the current training stage and to reduce the influence of large early losses:}
\begin{equation}
\mu_t=(1-\eta)\mu_{t-1}+\eta \ell_t,
\label{eq:ema_loss}
\end{equation}
where $\eta$ is the EMA update rate and $\ell_t$ is the training loss of the current sample. When computing the scaling factor, the historical baseline before the update, $\mu_{t-1}$, is compared with the current loss:
\begin{equation}
r_i=\begin{cases}
\max\left(\beta,\dfrac{\mu_{t-1}}{\ell_t+\epsilon}\right),
& \begin{array}{l}
s_i=\mathrm{internal}, \\[-1mm]
\end{array}\\[3mm]
1, & \text{otherwise}.
\end{cases}
\label{eq:kbm_scale}
\end{equation}
For samples where internal knowledge is correct but external retrieval is wrong, this scaling moderately reduces overly large penalties and protects the model's ability to answer with reliable parametric knowledge. The lower bound $\beta$ prevents such samples from being completely ignored. For other source labels, $r_i=1$, so no adjustment is applied.

Reliable-supervision records use only the CE term, while conflict-oriented records use joint positive-negative fine-tuning. The main source-aware objective is:
\begin{equation}
\begin{aligned}
\mathcal{L}_{\mathrm{main}}
&=\frac{1}{N}\sum_{i=1}^{N}\omega_i\big[
\mathcal{L}_{\mathrm{CE}}(x_i,y_i^+) 
+\lambda_{\mathrm{neg}}\mathbb{I}(y_i^-)
\mathcal{L}_{\mathrm{UL}}(x_i,y_i^-)\big],
\end{aligned}
\label{eq:main_objective}
\end{equation}
where $\mathbb{I}(y_i^-)$ indicates whether the $i$-th record contains an incorrect candidate, $\lambda_{\mathrm{neg}}$ controls the suppression strength, and $\omega_i$ is the sample-level weight:
\begin{equation}
\omega_i=w_{\tau(i)}\cdot \gamma_{s_i}\cdot r_i,
\label{eq:sample_weight}
\end{equation}
where $w_{\tau(i)}$ is determined by the functional type, $\gamma_{s_i}$ is determined by the knowledge-source label, and $r_i$ is the stabilization factor for internal-correct/external-wrong samples.

\subsection{Answer Completeness Regularization}

\subsubsection{Motivation: Incomplete Answer Generation}
Source-aware fine-tuning improves knowledge selection, but it may also exacerbate answer-incompleteness errors. In these cases, the model often starts to generate the correct answer but stops at a partial word, a partial entity, or the first fragment of a multi-word answer. Such outputs are especially harmful under exact-match evaluation and also reduce semantic clarity for downstream users.

\begin{table}[t]
\centering
\caption{Examples of incomplete answers without answer completeness regularization.}
\label{tab:incomplete_examples}
\scriptsize
\setlength{\tabcolsep}{4pt}
\renewcommand{\arraystretch}{1.15}
\begin{tabular}{p{0.52\columnwidth}p{0.18\columnwidth}p{0.22\columnwidth}}
\toprule
\textbf{Question} & \textbf{Incomplete output} & \textbf{Complete answer} \\
\midrule
What position does the spouse of the head of state of the country where Heath Ledger is a citizen hold? &  Consor & Consort of the United Kingdom \\
What position is held by the spouse of the head of government of the country where John F. Kelly is a citizen? & First & First Lady of the United States \\
What is the historical significance of Fort Dearborn in relation to the creator of \emph{The Mandalorian}? & Fort Dear & Fort Dearborn \\
\bottomrule
\end{tabular}
\vspace{-0.5cm}
\end{table}

Table~\ref{tab:incomplete_examples} shows that these errors are not always caused by choosing the wrong knowledge source. Outputs such as ``Consor'', ``Fort Dear'', and ``First'' suggest that the model has identified the correct answer direction but terminates before completing the answer. Therefore, TRACE augments source-aware fine-tuning with an auxiliary objective that explicitly encourages complete final-answer generation.

\subsubsection{Design of Answer Completeness Regularization}
Answer completeness regularization has two terms: answer-tail token reinforcement and premature termination suppression. The first term gives stronger supervision to later answer tokens, while the second term discourages termination tokens before the answer span is complete.

First, answer-tail token reinforcement strengthens the learning signal inside the answer span. Let the target output sequence be $y=\{y_1,y_2,\ldots,y_T\}$, where the answer content is located in \texttt{\textless answer\textgreater...\textless/answer\textgreater}. Let $m_t\in\{0,1\}$ indicate whether the $t$-th token is inside the answer span, and let $a_t\in[0,1]$ denote the relative position of this token inside the answer span. A larger $a_t$ means that the token is closer to the end of the answer. The coefficient $\rho$ controls the magnitude of answer-tail reinforcement. For a single answer, the answer-span weighted supervision loss is defined as:
\begin{equation}
\begin{aligned}
\mathcal{L}_{\mathrm{ans}}
&=-\frac{1}{\sum_{t=1}^{T}m_t+\epsilon}
\sum_{t=1}^{T}m_t(1+\rho a_t) \log p_{\theta}(y_t\mid x,y_{<t}).
\end{aligned}
\label{eq:answer_tail_loss}
\end{equation}
This means that in a long answer, later tokens receive relatively larger weights than earlier tokens. The design prevents the model from generating only the answer prefix while ignoring the latter part of multi-word entities, restrictive phrases, or long proper names, thereby reducing cases such as generating ``Consor'' instead of ``Consort of the United Kingdom''.

Second, premature termination suppression reduces early stopping before the answer is complete. Let $E$ denote the set of termination tokens, including EOS and \texttt{\textless/answer\textgreater} related tokens. Inside the answer-content span, the model should continue generating answer content, so the probability of termination should be reduced. Let $p_t^{\mathrm{end}}$ be the termination probability at the $t$-th position:
\begin{equation}
p_t^{\mathrm{end}}=\sum_{e\in E}p_{\theta}(e\mid x,y_{<t}).
\label{eq:end_prob}
\end{equation}
The premature termination suppression loss is:
\begin{equation}
\mathcal{L}_{\mathrm{end}}=-\frac{1}{\sum_{t=1}^{T}m_t+\epsilon}\sum_{t=1}^{T}m_t\log(1-p_t^{\mathrm{end}}).
\label{eq:end_loss}
\end{equation}
This term is essentially an unlikelihood constraint on termination tokens. It suppresses early generation of ending tags before the answer body is complete, reducing empty answers, overly short answers, truncated entities, and incomplete answer boundaries.

The answer completeness regularization for a single answer is:
\begin{equation}
\mathcal{L}_{\mathrm{comp}}=\mathcal{L}_{\mathrm{ans}}+\lambda_{\mathrm{end}}\mathcal{L}_{\mathrm{end}}.
\label{eq:comp_loss}
\end{equation}

The final training objective combines source-aware fine-tuning and answer completeness regularization:
\begin{equation}
\mathcal{L}_{\mathrm{total}}=\mathcal{L}_{\mathrm{main}}+\lambda_{\mathrm{comp}}\mathcal{L}_{\mathrm{comp}},
\label{eq:total_loss}
\end{equation}
where $\lambda_{\mathrm{comp}}$ controls the contribution of answer completeness regularization.

\subsubsection{Difference from Format Constraints}
Answer completeness regularization is different from a pure format constraint. A format constraint asks whether the answer is placed in the required template and whether tags are well formed. For example, \texttt{\textless answer\textgreater...\textless/answ} is a malformed tag and belongs to the scope of format constraints. In contrast, answer completeness regularization asks whether the content inside a valid answer span is semantically complete.

For example, the output ``First'' can be placed inside a valid answer tag, but it is incomplete when the golden answer is ``First Lady of the United States.'' TRACE therefore treats answer completeness as an independent generation objective. It does not require the model to generate longer answers in general; rather, it encourages the model to preserve complete entities, restrictive phrases, and answer boundaries once the final knowledge source has been selected.

\section{Experiments and Results}

We evaluate around four questions that directly correspond to the claims in the preceding sections. \textbf{RQ1} asks whether TRACE can preserve the ability to use correct retrieved context while resisting misleading retrieved context. \textbf{RQ2} asks whether the same source-aware training remains effective under more complex retrieval conditions, including self-conflicting, irrelevant, and partially relevant contexts. \textbf{RQ3} asks whether debate traces, strong-to-weak supervision, and answer completeness regularization are necessary. \textbf{RQ4} asks whether the debate-trace training framework can improve a substantially weaker target model.

\subsection{Experimental Setup}
The evaluation covers five retrieved-context scenarios. \textbf{Scenario I (S1)} uses correct contextual knowledge, where retrieved passages support the golden answer. \textbf{Scenario II (S2)} uses adversarial contextual knowledge, where retrieved passages contradict the golden answer. We further evaluate three harder retrieval settings: \textbf{Scenario III (S3)}, self-conflicting contextual knowledge; \textbf{Scenario IV (S4)}, irrelevant contextual knowledge; and \textbf{Scenario V (S5)}, partially relevant contextual knowledge mixed with distractors.

\subsubsection{Models and Baselines}
We use Qwen2.5--7B-Instruct and Llama3.1--8B-Instruct as the main backbone models to evaluate whether TRACE is stable across model families. In the debate stage, unless otherwise specified, the critic model is Qwen3--32B and the judge model is DeepSeek-R1--8B. The maximum number of debate rounds is 5. However, the debate can terminate earlier based on the judge model’s stopping decision, resulting in fewer than 5 rounds per question on average.

We compare TRACE with query-only prompting, RAG prompting, Astute-RAG~\cite{wang2025astute}, CK-PlUG~\cite{bi2026parameters}, supervised fine-tuning (SFT), GRPO with RAG~\cite{guo2025deepseek}, and Knowledgeable-R1~\cite{lin2026resisting}. These baselines cover no-retrieval prompting, standard retrieval prompting, prompt-based conflict handling, learnable knowledge-source control, supervised fine-tuning, and reinforcement-learning-based source selection. We use exact match (EM) as the primary metric, following the standard protocol of these benchmarks, and report all scores as percentages where higher is better. All fine-tuned variants are trained for one epoch. During inference, all methods use deterministic decoding with temperature set to 0 to reduce sampling variance.

TRACE is implemented with LoRA fine-tuning~\cite{hu2022lora}, while some compared methods use full-parameter reinforcement-learning fine-tuning and report results on H100 GPUs. Our experiments are conducted on RTX 4090 GPUs. This setting emphasizes whether debate-trace supervision can provide practical gains under a parameter-efficient training regime.

\subsubsection{Datasets}
The datasets cover correct, wrong, self-conflicting, irrelevant, and partially relevant retrieval scenarios. S1 and S2 are mainly constructed from ConFiQA~\cite{bi2026context} and include three subtasks: QA, MR, and MC. QA denotes single-hop question answering. MR denotes multi-hop reasoning where only one evidence step is wrong. MC denotes multi-hop reasoning where multiple steps in the evidence chain are wrong. PC-QA, PC-MR, and PC-MC evaluate the model's ability to use correct retrieved knowledge, whereas NC-QA, NC-MR, and NC-MC evaluate robustness when the retrieved context contradicts the correct answer.

For extended experiments, S3 uses the SC dataset constructed from ConFiQA, where mutually conflicting evidence is provided to evaluate the model's ability to handle internal contradictions within the context. S4 uses the ExplainPE medical question-answering dataset~\cite{wen2024mindmap}, where the retrieved content is largely irrelevant to the question, evaluating the model's resistance to irrelevant-context interference. S5 uses HotPotQA~\cite{yang2018hotpotqa}, 2WikiMultiHopQA~\cite{ho2020constructing}, and MuSiQue~\cite{trivedi2022musique}, where useful evidence and distracting passages appear together, evaluating evidence selection under partially relevant retrieval results.

\subsubsection{Source of Retrieved Contexts}
For all datasets, we use the contexts provided by the benchmarks and do not introduce an additional retriever. Correct and wrong passages are directly taken from the benchmark structure rather than re-indexed or re-retrieved from a larger corpus. We use the first five passages provided by each benchmark and keep their original order. Truncation is applied only when the input exceeds the maximum length supported by the backbone model.

\subsection{Results on Standard Conflict Scenarios}
The standard conflict scenarios evaluate the central requirement of source-aware RAG: using reliable context in S1 while resisting wrong context in S2. Table~\ref{tab:main_results} reports EM results on both backbone models.

\paragraph{Scenario I: Correct contextual knowledge (S1)}
TRACE preserves strong context utilization when retrieved evidence is reliable. On Qwen2.5--7B-Instruct, TRACE achieves 90.92\%, 91.05\%, and 86.91\% on PC-MR, PC-MC, and PC-QA, respectively, outperforming Knowledgeable-R1 by 15.84, 15.54, and 6.01 percentage points. On Llama3.1--8B-Instruct, TRACE reaches 91.41\%, 84.63\%, and 91.91\%, improving over Knowledgeable-R1 by 17.65, 4.39, and 11.88 percentage points. These results show that TRACE does not gain robustness by simply rejecting retrieved content; it continues to exploit correct external evidence.

\paragraph{Scenario II: Adversarial contextual knowledge (S2)}
TRACE also improves robustness when retrieved evidence is wrong, although the margin depends on the backbone and subtask. Conventional RAG prompting suffers a clear drop under S2; for example, on Qwen2.5--7B-Instruct, it obtains only 13.47\%, 8.06\%, and 11.31\% on NC-MR, NC-MC, and NC-QA, all lower than query-only prompting. In contrast, TRACE reaches 47.64\%, 39.04\%, and 37.48\%, improving over GRPO w/ RAG by 20.70, 19.30, and 11.47 percentage points and over Knowledgeable-R1 by 3.70, 1.70, and 8.08 percentage points. On Llama3.1--8B-Instruct, TRACE reaches 50.17\%, 43.42\%, and 44.91\%. It is best on NC-MC and NC-QA, while Knowledgeable-R1 remains stronger on NC-MR. Overall, the S2 results support the main claim that debate-trace supervision improves resistance to misleading retrieved content without removing the ability to use correct context.

\begin{table}[t]
\centering
\caption{EM (\%) in standard conflict scenarios. The best result is in bold, and the second-best result is underlined.}
\label{tab:main_results}
\scriptsize
\setlength{\tabcolsep}{3.8pt}
\renewcommand{\arraystretch}{1.08}
\resizebox{\columnwidth}{!}{%
\begin{tabular}{lcccccc}
\toprule
 & \multicolumn{3}{c}{\textbf{S1: Correct}} & \multicolumn{3}{c}{\textbf{S2: Wrong}} \\
\cmidrule(lr){2-4}\cmidrule(lr){5-7}
\textbf{Method} & \textbf{PC-MR} & \textbf{PC-MC} & \textbf{PC-QA} & \textbf{NC-MR} & \textbf{NC-MC} & \textbf{NC-QA} \\
\midrule
\multicolumn{7}{l}{\textit{Qwen2.5--7B}} \\
Query-only prompting & 27.72\% & 24.66\% & 31.67\% & 25.93\% & 25.82\% & \second{32.31\%} \\
RAG prompting & 65.68\% & 66.39\% & 74.35\% & 13.47\% & 8.06\% & 11.31\% \\
CK-PlUG & 64.69\% & 66.55\% & 78.66\% & 11.62\% & 8.06\% & 7.92\% \\
Astute-RAG & 65.51\% & 66.05\% & 77.62\% & 12.79\% & 7.07\% & 10.34\% \\
SFT & 71.95\% & \second{77.70\%} & 74.70\% & 24.92\% & 21.05\% & 21.97\% \\
GRPO w/ RAG & \second{77.56\%} & 77.36\% & 80.03\% & 26.94\% & 19.74\% & 26.01\% \\
Knowledgeable-R1 & 75.08\% & 75.51\% & \second{80.90\%} & \second{43.94\%} & \second{37.34\%} & 29.40\% \\
TRACE & \best{90.92\%} & \best{91.05\%} & \best{86.91\%} & \best{47.64\%} & \best{39.04\%} & \best{37.48\%} \\
\midrule
\multicolumn{7}{l}{\textit{Llama3.1--8B}} \\
Query-only prompting & 29.37\% & 26.18\% & 39.93\% & 27.10\% & 27.63\% & 42.65\% \\
RAG prompting & 64.85\% & 61.99\% & 76.42\% & 22.90\% & 16.28\% & 24.88\% \\
CK-PlUG & 54.79\% & 58.45\% & 69.71\% & 12.12\% & 9.05\% & 17.29\% \\
Astute-RAG & 65.84\% & 64.86\% & 77.97\% & 17.00\% & 9.05\% & 17.29\% \\
SFT & 72.88\% & 79.22\% & 73.84\% & 42.59\% & 35.53\% & 35.86\% \\
GRPO w/ RAG & \second{78.05\%} & 79.73\% & \second{82.62\%} & 41.58\% & 35.69\% & 39.26\% \\
Knowledgeable-R1 & 73.76\% & \second{80.24\%} & 80.03\% & \best{55.39\%} & \second{41.12\%} & \second{44.59\%} \\
TRACE & \best{91.41\%} & \best{84.63\%} & \best{91.91\%} & \second{50.17\%} & \best{43.42\%} & \best{44.91\%} \\
\bottomrule
\end{tabular}%
}
\vspace{-0.5cm}
\end{table}

\subsection{Extended Evaluation under More Complex Retrieval Scenarios}
The extended evaluation tests whether TRACE generalizes beyond the standard correct-context and wrong-context setting. Table~\ref{tab:extended_results} reports EM results under self-conflicting, irrelevant, and partially relevant retrieval.

\paragraph{Scenario III: Self-conflicting contextual knowledge (S3)}
TRACE is particularly effective when retrieved evidence is internally contradictory. In S3, the model must choose among mutually conflicting contextual claims rather than merely decide whether to use retrieval. TRACE reaches 90.58\% on Qwen2.5--7B-Instruct, outperforming Knowledgeable-R1 by 14.25 percentage points, and reaches 92.58\% on Llama3.1--8B-Instruct, outperforming GRPO w/ RAG by 16.00 percentage points. This result is consistent with the method design: debate traces expose competing candidate answers and answer shifts, which provide direct supervision for resolving self-conflicting evidence.

\paragraph{Scenario IV: Irrelevant contextual knowledge (S4)}
TRACE remains competitive when the retrieved context is largely irrelevant, but the strength of the result is backbone-dependent. On ExplainPE with Qwen2.5--7B-Instruct, query-only prompting obtains 64.45\%, while RAG prompting drops to 62.21\%, showing that irrelevant retrieval can hurt answer accuracy. TRACE reaches 69.14\%, outperforming Knowledgeable-R1 and indicating stronger resistance to irrelevant-context interference. On Llama3.1--8B-Instruct, TRACE reaches 47.95\%, improving over RAG prompting by 8.79 percentage points and slightly outperforming GRPO w/ RAG, but remaining below Knowledgeable-R1.

\paragraph{Scenario V: Partially relevant contextual knowledge (S5)}
S5 reveals the main boundary of TRACE. On Qwen2.5--7B-Instruct, TRACE is best on MuSiQue and second-best on 2Wiki, but it is below Knowledgeable-R1 on HotPotQA. On Llama3.1--8B-Instruct, TRACE is second-best on MuSiQue but lags behind the strongest baseline on HotPotQA and 2Wiki. This pattern suggests that TRACE is strongest when the central challenge is source conflict or misleading candidates, while partially relevant long-context multi-hop settings require finer paragraph-level evidence localization and evidence-chain composition. These results motivate future extensions that combine debate-trace supervision with explicit evidence-chain supervision.

\begin{table}[t]
\centering
\caption{EM (\%) in more complex retrieval scenarios. The best result is in bold, and the second-best result is underlined.}
\label{tab:extended_results}
\scriptsize
\setlength{\tabcolsep}{4pt}
\renewcommand{\arraystretch}{1.08}
\resizebox{\columnwidth}{!}{%
\begin{tabular}{lccccc}
\toprule
 & \textbf{S3: Conflict} & \textbf{S4: Irrelevant} & \multicolumn{3}{c}{\textbf{S5: Partly Irrelevant}} \\
\cmidrule(lr){2-2}\cmidrule(lr){3-3}\cmidrule(lr){4-6}
\textbf{Method} & \textbf{SC} & \textbf{ExplainPE} & \textbf{HotPotQA} & \textbf{2Wiki} & \textbf{MuSiQue} \\
\midrule
\multicolumn{6}{l}{\textit{Qwen2.5--7B}} \\
Query-only prompting & 29.67\% & 64.45\% & 20.90\% & 25.54\% & 4.36\% \\
RAG prompting & 59.50\% & 62.21\% & 20.36\% & 22.53\% & 6.41\% \\
CK-PlUG & 55.00\% & 55.00\% & 22.74\% & 24.76\% & 6.25\% \\
Astute-RAG & 54.20\% & 56.74\% & 17.87\% & 20.35\% & 6.29\% \\
SFT & 68.50\% & 66.60\% & \second{30.14\%} & 32.20\% & 11.75\% \\
GRPO w/ RAG & 75.33\% & 66.50\% & 27.93\% & 33.95\% & 11.79\% \\
Knowledgeable-R1 & \second{76.33\%} & \second{67.57\%} & \best{31.45\%} & \best{37.52\%} & \second{12.04\%} \\
TRACE & \best{90.58\%} & \best{69.14\%} & 29.29\% & \second{37.44\%} & \best{12.37\%} \\
\midrule
\multicolumn{6}{l}{\textit{Llama3.1--8B}} \\
Query-only prompting & 32.08\% & 43.26\% & 20.69\% & 21.02\% & 6.16\% \\
RAG prompting & 61.17\% & 39.16\% & 24.44\% & 23.50\% & 8.19\% \\
CK-PlUG & 42.00\% & 31.54\% & 22.35\% & 24.63\% & 5.25\% \\
Astute-RAG & 59.83\% & 40.14\% & 1.65\% & 30.26\% & 9.64\% \\
SFT & 70.12\% & 47.17\% & 33.59\% & 38.24\% & 13.36\% \\
GRPO w/ RAG & \second{76.58\%} & 47.56\% & \second{34.84\%} & \second{41.22\%} & \best{16.59\%} \\
Knowledgeable-R1 & 73.67\% & \best{49.61\%} & \best{37.06\%} & \best{45.37\%} & 14.69\% \\
TRACE & \best{92.58\%} & \second{47.95\%} & 31.17\% & 38.88\% & \second{15.18\%} \\
\bottomrule
\end{tabular}%
}
\end{table}

\subsection{Ablation Study}
The ablation study evaluates whether the main modules in TRACE are responsible for the observed gains. We evaluate three variants on the QA, MC, and MR subsets of ConFiQA. \emph{Without debate} removes multi-agent debate and directly uses query-only and RAG responses from the base LLM. \emph{Without strong-teaching-weak} keeps the debate format but replaces all debate roles with the base LLM. \emph{Without answer completeness regularization} sets the weight of answer completeness regularization to 0.

\begin{table}[t]
\centering
\caption{Ablation results on ConFiQA EM (\%). The best result in each column is in bold.}
\label{tab:ablation}
\scriptsize
\setlength{\tabcolsep}{5pt}
\renewcommand{\arraystretch}{1.15}
\resizebox{\columnwidth}{!}{%
\begin{tabular}{lcccccc}

\toprule
\textbf{Variant} & \textbf{PC-MC} & \textbf{NC-MC} & \textbf{PC-MR} & \textbf{NC-MR} & \textbf{PC-QA} & \textbf{NC-QA} \\
\midrule
Full TRACE & 91.05\% & \best{39.04\%} & \best{90.92\%} & \best{47.64\%} & 86.91\% & \best{37.48\%} \\
w/o debate & 85.14\% & 35.86\% & 86.30\% & 44.94\% & \best{87.26\%} & 36.99\% \\
w/o strong-teaching-weak & \best{92.06\%} & 38.15\% & 87.45\% & 45.45\% & 85.54\% & 35.86\% \\
\makecell{w/o answer completeness  regularization} &
85.97\% & 38.65\% & 83.00\% & 42.59\% & 85.88\% & 32.14\% \\
\bottomrule
\end{tabular}%
}
\vspace{-0.5cm}
\end{table}

Removing debate reduces the average EM from 65.51\% to 62.75\%, with drops on all wrong-context metrics. The only local improvement is PC-QA, which increases from 86.91\% to 87.26\%. This pattern indicates that simple single-hop questions under correct context can be solved without debate traces, but robustness under misleading retrieval benefits from intermediate candidate answers and positive-negative samples exposed by debate.

Replacing strong-to-weak debate with same-model debate also weakens overall performance. PC-MC increases from 91.05\% to 92.06\%, but NC-MC, PC-MR, NC-MR, PC-QA, and NC-QA all decrease, and average EM drops from 65.51\% to 64.09\%. This result suggests that stronger critic and judge models provide useful correction and counterfactual-discrimination signals, especially under wrong-context settings.

Removing answer completeness regularization causes the largest average drop, from 65.51\% to 61.37\%. All six metrics decrease, with the largest drop on NC-QA. This supports the claim that source selection alone is insufficient: after the model identifies a reliable answer direction, answer-tail reinforcement and premature-termination suppression help preserve complete answer boundaries and reduce incomplete outputs.

To complement these quantitative ablations, Appendix~\ref{app:case_analysis} provides two representative cases that illustrate how TRACE resists misleading retrieved evidence and preserves complete answer boundaries.

\subsection{Role Conversion between Strong and Weak Models}
The role-conversion experiment tests whether TRACE can transfer supervision from relatively stronger models to a substantially weaker target model. In the main setting, the strong model is Qwen3--32B, the intermediate model is DeepSeek-R1--8B, and the weak target model is Qwen2.5--7B. In the role-conversion setting, Qwen2.5--7B is used as the strongest model, Qwen2.5--3B is used as the intermediate model, and Qwen2.5--1.5B is used as the weak target model for fine-tuning.

\begin{figure}[t]
\centerline{\includegraphics[width=0.85\linewidth,trim=0 20 0 0]{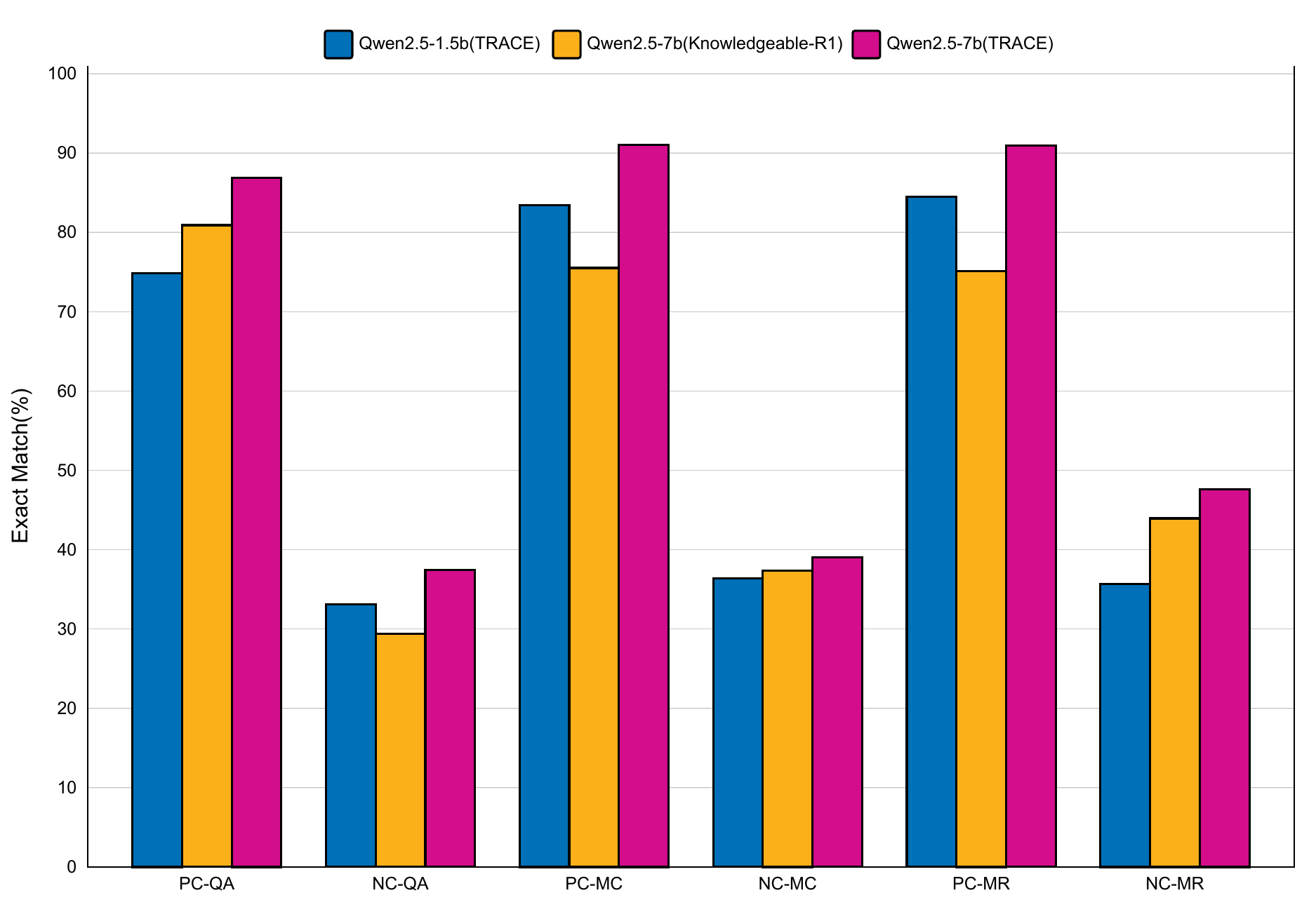}}
\caption{Role-conversion results between stronger debate models and a weaker target model, with comparison to the strongest baseline.}
\label{fig:role_conversion}
\vspace{-0.5cm}
\end{figure}

Fig.~\ref{fig:role_conversion} shows that TRACE can still improve a weak target model in several knowledge-conflict settings. Although Qwen2.5--1.5B has only 21.3\% of the parameters of Qwen2.5--7B, the fine-tuned model remains competitive and surpasses the 7B-level Knowledgeable-R1 baseline on some metrics. This result suggests that debate-trace supervision can transfer useful source-selection behavior to weaker models, although the effect is not uniform across all scenarios.

\section{Conclusion and Future Work}
This paper addresses knowledge conflicts in RAG by proposing TRACE, a source-aware fine-tuning framework that learns when to use retrieved context and when to rely on parametric knowledge. The key idea is to use multi-agent debate only during data construction: debate trajectories expose correct candidates, misleading candidates, and answer-shift patterns, which are converted into positive and negative fine-tuning signals for a single target model. TRACE further adds answer completeness regularization so that source selection is paired with complete final-answer generation.

Experiments across correct, wrong, self-conflicting, irrelevant, and partially relevant retrieval settings show that TRACE preserves the use of reliable context while improving robustness to misleading or conflicting retrieval. Ablation results confirm the contribution of debate traces, strong-to-weak supervision, and answer completeness regularization, and the role-conversion study suggests that the supervision can also benefit a weaker target model. These findings indicate that debate trajectories can serve as process-level supervision for source-aware RAG without requiring debate at inference time.

A current limitation is that TRACE is strongest in explicit source-conflict settings and less uniformly effective in partially relevant long-context multi-hop scenarios, where answer correctness depends on fine-grained evidence localization and evidence-chain composition. Future work will therefore combine debate-trace supervision with explicit evidence-chain supervision, reduce the cost and noise of debate-based data construction, and evaluate source-aware training under more realistic retrieval pipelines beyond benchmark-provided contexts.

\section*{Acknowledgments}
This research was supported by the Key R\&D Program of Shandong Province, China (Project No. 2024CXGC010109), the Shandong Provincial Natural Science Foundation (Project No. ZR2026QC1574), and the 2026 Linyi University High-Level Talent (Doctoral) Research Start-up Fund (Natural Science), Grant No. Z6126063.

\bibliography{reference}
\appendices
\section{Case Analysis}
\label{app:case_analysis}

This appendix provides two representative cases to illustrate the behavior of TRACE under misleading retrieved contexts. We summarize the key misleading evidence in each case and compare the outputs of Knowledgeable-R1~\cite{lin2026resisting} and TRACE. Case 1 mainly reflects context resistance under erroneous retrieval, and case 2 shows the importance of answer completeness regularization mechanism.

\subsection{Case 1: Resisting a Misleading Citizenship Chain}
\label{app:case_robert_evans}

\noindent\textbf{Question.}
\begin{quote}
What is the native language of the head of government of the country where Robert Evans has citizenship?
\end{quote}

\noindent\textbf{Misleading evidence.}
The retrieved passage incorrectly describes Robert Evans as a citizen of Belgium. It then builds a misleading reasoning chain from Belgium to its head of government and further associates the final answer with Dutch. The passage also contains noisy and internally inconsistent statements, such as connecting Belgium with Narendra Modi and describing him as a native Puerto Rican Spanish speaker.

\noindent\textbf{Knowledgeable-R1 output.}
\begin{quote}
\footnotesize\ttfamily\raggedright
\textless think\textgreater{} Robert Evans is a citizen of Belgium. The head of government in Belgium is the Prime Minister. The native language of the Prime Minister of Belgium is Dutch. \textless /think\textgreater{}

\textless answer\textgreater{} Dutch \textless /answer\textgreater{}
\end{quote}

\noindent\textbf{TRACE output.}
\begin{quote}
\footnotesize\ttfamily\raggedright
\textless think\textgreater{} The information provided about Robert Evans is incorrect, as he is a U.S. citizen, not a Belgian. However, the question asks about his citizenship, which is the United States. The head of government of the United States is the President, and the native language of the President is English. \textless /think\textgreater{}

\textless answer\textgreater{} English \textless /answer\textgreater{}
\end{quote}

\noindent\textbf{Analysis.}
This case illustrates the context-resistance ability of TRACE. The thinking process of Knowledgeable-R1 shows that it accepts the incorrect retrieved premise that Robert Evans is a Belgian citizen, and then follows the misleading chain ``Belgium, Prime Minister,Dutch.'' As a result, it produces an incorrect final answer. By contrast, the thinking process of TRACE explicitly identifies the retrieved citizenship statement as unreliable and falls back to parametric knowledge. It then maps the country to the United States, the head of government to the U.S. President, and the native language to English. This example provides qualitative evidence that debate-trace supervision can help the model resist misleading retrieved content when parametric knowledge is more reliable.
\subsection{Case 2: Preserving the Complete Answer Boundary}
\label{app:case_bam_adebayo}

\noindent\textbf{Question.}
\begin{quote}
What position is held by the spouse of the head of state of the country where Bam Adebayo is a citizen?
\end{quote}

\noindent\textbf{Misleading evidence.}
The retrieved passage contains noisy and irrelevant descriptions. It incorrectly introduces Bam Adebayo as being from the Russian Empire, then shifts to unrelated statements about Charles's Empire, Charles III, and Ronald Reagan. 

\noindent\textbf{Model outputs.}
Knowledgeable-R1 produces an answer in the correct general direction but with an incomplete answer:
\begin{quote}
\texttt{<answer> First Lady </answer>}
\end{quote}
TRACE gives a more complete answer:
\begin{quote}
\texttt{<answer> First Lady of the United States </answer>}
\end{quote}

\noindent\textbf{Analysis.}
The core of this analysis is not mainly about choosing between parametric knowledge or retrieved knowledge. The key difference lies in answer completeness. Knowledgeable-R1 outputs only the head noun phrase, while TRACE preserves the complete answer boundary by including the country-specific qualifier ``of the United States.'' For exact-match style evaluation and for semantic clarity, this qualifier is important because it makes the final answer more specific and less ambiguous. This example therefore supports the role of answer-completeness regularization: answer-tail token reinforcement and premature termination suppression help LLM generate a complete final answer rather than stopping at a shorter answer fragment.

\subsection{Summary}
\label{app:case_summary}

The first case shows that TRACE can resist some misleading retrieved chain and recover the answer from reliable parametric knowledge more effectively during some situations. The second case shows that TRACE can produce a more complete answer. Together, they provide qualitative evidence for the two central components of TRACE: the fine-tuning method based on multi-agent debate traces helps LLM choose more reliable knowledge and answer completeness regularization mechanism helps LLMs reduce the generation of incomplete answer.
\end{document}